\documentclass[10pt,twocolumn,letterpaper]{article}

\usepackage[dvipsnames]{xcolor}
\usepackage{wacv}
\usepackage{times}
\usepackage{epsfig}
\usepackage{graphicx}
\usepackage{amsmath}
\usepackage{amssymb}
\usepackage{booktabs}
\usepackage{multirow}

\usepackage[sort]{cite}
\usepackage{enumitem}
\usepackage{placeins}
\usepackage[accsupp]{axessibility}

\usepackage{xcolor}
\PassOptionsToPackage{hyphens}{url}\usepackage[pagebackref=true,breaklinks=true,colorlinks,citecolor=citecolor,bookmarks=false]{hyperref}
\definecolor{citecolor}{RGB}{34,139,34}
\definecolor{myred}{RGB}{206,45,31}
\definecolor{lightcoral}{RGB}{225,134,131}
\definecolor{lightgreen}{RGB}{166,236,153}

\newcommand{\red}[1]{\textcolor{red}{#1}}
\newcommand{\green}[1]{\textcolor{OliveGreen}{#1}}

\makeatletter
\newcommand\notsotiny{\@setfontsize\notsotiny\@vipt\@viipt}
\makeatother

\wacvfinalcopy 
\ifwacvfinal
\usepackage[breaklinks=true,bookmarks=false]{hyperref}
\else
\usepackage[pagebackref=true,breaklinks=true,colorlinks,bookmarks=false]{hyperref}
\fi

\begin{document}

\title{Do Pre-trained Models Benefit Equally in Continual Learning?}

\author{
Kuan-Ying Lee \and Yuanyi Zhong \and Yu-Xiong Wang \and \\ University of Illinois at Urbana-Champaign \\ {\tt\small \{kylee5, yuanyiz2, yxw\}@illinois.edu}
}

\maketitle
\thispagestyle{empty}

\begin{abstract}
Existing work on continual learning (CL) is primarily devoted to developing algorithms for models trained from scratch. Despite their encouraging performance on contrived benchmarks, these algorithms show dramatic performance drop in real-world scenarios.
Therefore, this paper advocates the systematic introduction of pre-training to CL, which is a general recipe for transferring knowledge to downstream tasks but is substantially missing in the CL community.
Our investigation reveals the multifaceted complexity of exploiting pre-trained models for CL, along three different axes: pre-trained models, CL algorithms, and CL scenarios.
Perhaps most intriguingly, improvements in CL algorithms from pre-training are very inconsistent -- an underperforming algorithm could become competitive and even state of the art, when all algorithms start from a pre-trained model.
This indicates that the current paradigm, where all CL methods are compared in from-scratch training, is not well reflective of the true CL objective and desired progress.
In addition, we make several other important observations, including that 1) CL algorithms that exert less regularization benefit more from a pre-trained model; and 2) a stronger pre-trained model such as CLIP does not guarantee a better improvement. 
Based on these findings, we introduce a simple yet effective baseline that employs minimum regularization and leverages the more beneficial pre-trained model, coupled with a two-stage training pipeline. We recommend including this strong baseline in the future development of CL algorithms, due to its demonstrated state-of-the-art performance.
Our code is available at \url{https://github.com/eric11220/pretrained-models-in-CL}.
\end{abstract}

\vspace{-0.5em}
\section{Introduction} \label{sec:intro}
\begin{figure}
    \centering
    \includegraphics[width=\linewidth]{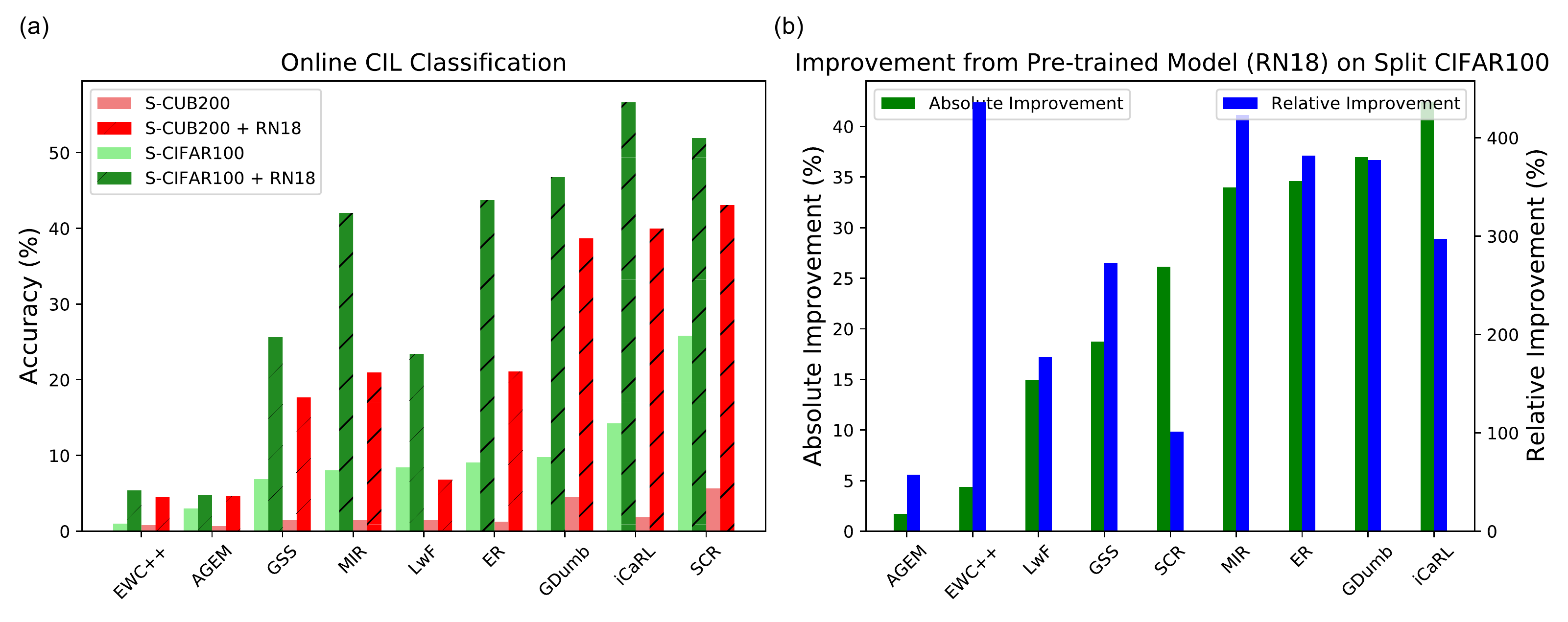}
    \caption{(a) CL algorithms trained from scratch fail on \textcolor{lightcoral}{Split CUB200}, a more complex dataset than \textcolor{lightgreen}{Split CIFAR100}, which necessitates the use of pre-trained models (denoted as `+ RN18') that dramatically increase the accuracy of a wide spectrum of algorithms.
    (b) Different CL algorithms receive vastly different benefits from pre-trained models, and the superiority between algorithms changes. These findings suggest that it is critical for the community to develop CL algorithms with a pre-trained model and understand their behaviors.
    \textbf{[Best viewed in color.]}} 
    \label{fig:teaser}
    \vspace{-1em}
\end{figure}

Continual learning (CL) has gained increasing research momentum recently, due to the ever-changing nature of real-world data \cite{cope,cndpm,maximally,icarl,cong2020gan,ewc,si,Aljundi_2018_ECCV,xu2018reinforced,rusu2016progressive}.
Despite their encouraging performance, many notable CL algorithms were developed to work with a model trained from scratch.
As one of the key objectives, this paper advocates the {\em systematic introduction of pre-training to CL}. This is rooted in the following {\em two} observed fundamental limitations of building CL algorithm on top of a from-scratch trained model, which fails to reflect the true progress in the CL research for real-world scenarios as shown in Fig.~\ref{fig:teaser}.

First, training from scratch does not reflect the actual performance, because if one were to apply a CL algorithm to real-world scenarios, it would be counter-intuitive not to build upon off-the-shelf pre-trained models given the large performance gap (Fig.~\ref{fig:teaser}).
One might argue that applying all algorithms to a from-scratch trained model simplifies comparison between different algorithms.
However, intriguingly, our study shows that {\em an underperforming algorithm could become competitive and even achieve state-of-the-art performance, when all algorithms start from a pre-trained model}.
In particular, iCaRL \cite{icarl}, which shows mediocre performance in online class incremental learning (CIL) when trained from scratch, is comparable to or even outperforms SCR~\cite{scr}, when both are initialized from a ResNet18\footnote{We refer to ResNet as RN throughout the paper.} pre-trained on ImageNet (accuracy increase from 14.26\% to 56.64\% for iCaRL vs. increase from 25.80\% to 51.93\% for SCR on Split CIFAR100 in Fig.~\ref{fig:teaser} and Table~\ref{tab:cifar100}).
This potentially indicates that the efforts funneled into the development of CL algorithms could be in a less effective direction and are not well reflective of the actual progress in CL.
Therefore, we should develop any future CL algorithms in the context of how we are going to use them in practice -- starting from a pre-trained model.

Second, for many more realistic datasets with diverse visual concepts, data scarcity makes it impossible to train a CL learner from scratch~\cite{agem} (as also shown in the results on Split CUB200 in Fig.~\ref{fig:teaser}. We believe that this is partially the reason why CL classification literature still heavily evaluates on contrived benchmarks such as Split MNIST and Split CIFAR~\cite{cope, cndpm}, as opposed to much more complex datasets typically used in offline learning.

Through our investigation, this paper reveals the {\em multifaceted complexity} of exploiting pre-trained models for CL.
As summarized in Table~\ref{tab:axes}, we conduct the investigation along {\em three} different axes: different pre-trained models, different CL algorithms, and different CL scenarios.
In particular, we analyze models pre-trained in either supervised or self-supervised fashion and from three distinct sources of supervision -- curated labeled images, non-curated image-text pairs, and unlabeled images. 
These models cover supervised RN18/50 \cite{he2016deep} trained on ImageNet classification \cite{deng2009imagenet}, CLIP RN50 \cite{radford2021learning}, and self-supervised RN50 trained with SimCLR \cite{simclr}, SwAV \cite{swav}, or Barlow Twins \cite{barlow_twins}.

We make several important observations.
\textbf{1)} Benefits of a pre-trained model on different CL algorithms vary widely, as represented by the aforementioned comparison between iCaRL and SCR.
\textbf{2)} As shown in Fig.~\ref{fig:teaser}, algorithms applying less regularization to the gradient (i.e., replay-based methods like ER~\cite{er}) seem to benefit the most from pre-trained models.
\textbf{3)} Intriguingly, despite its impressive zero-shot capability, CLIP RN50 mostly {\em underperforms} ImageNet RN50.
\textbf{4)} Self-supervised fine-tuning helps alleviate catastrophic forgetting. For example, fine-tuning SimCLR RN50 on the downstream dataset in a self-supervised fashion with the SimCLR loss demonstrates a huge reduction in forgetting, compared with supervised models (17.99\% forgetting of SimCLR RN50 vs. 91.12\% forgetting of supervised RN50). 
\textbf{5)} Iterating over data of a given task for multiple epochs as in class incremental learning (CIL) {\em does not necessarily} improve the performance over online CIL.

Based on these observations, we further propose a strong baseline by applying ER, which exerts minimum regularization (the second observation), on an ImageNet pre-trained model (the third observation).
Coupled with a two-stage training pipeline \cite{two_stage} (Sec.~\ref{subsec:two_stage}), we show that such a simple baseline achieves state-of-the-art performance.
We recommend including this strong baseline in the future development of CL algorithms.

{\bf Our contributions} are summarized as follows. 1) We show the necessity of pre-trained models on more complex CL datasets and the dramatic difference in their benefits on different CL algorithms, which may overturn the comparison results between algorithms. Therefore, we suggest the community consider pre-trained models when developing and evaluating new CL algorithms. 2) We show that replay-based CL algorithms seem to benefit more from a pre-trained model, compared with regularization-based counterparts. 3) We propose a simple yet strong baseline based on ER and ImageNet RN50, which achieves state-of-the-art performance for CL with pre-training.

\begin{table}
    \centering
    \includegraphics[width=\linewidth]{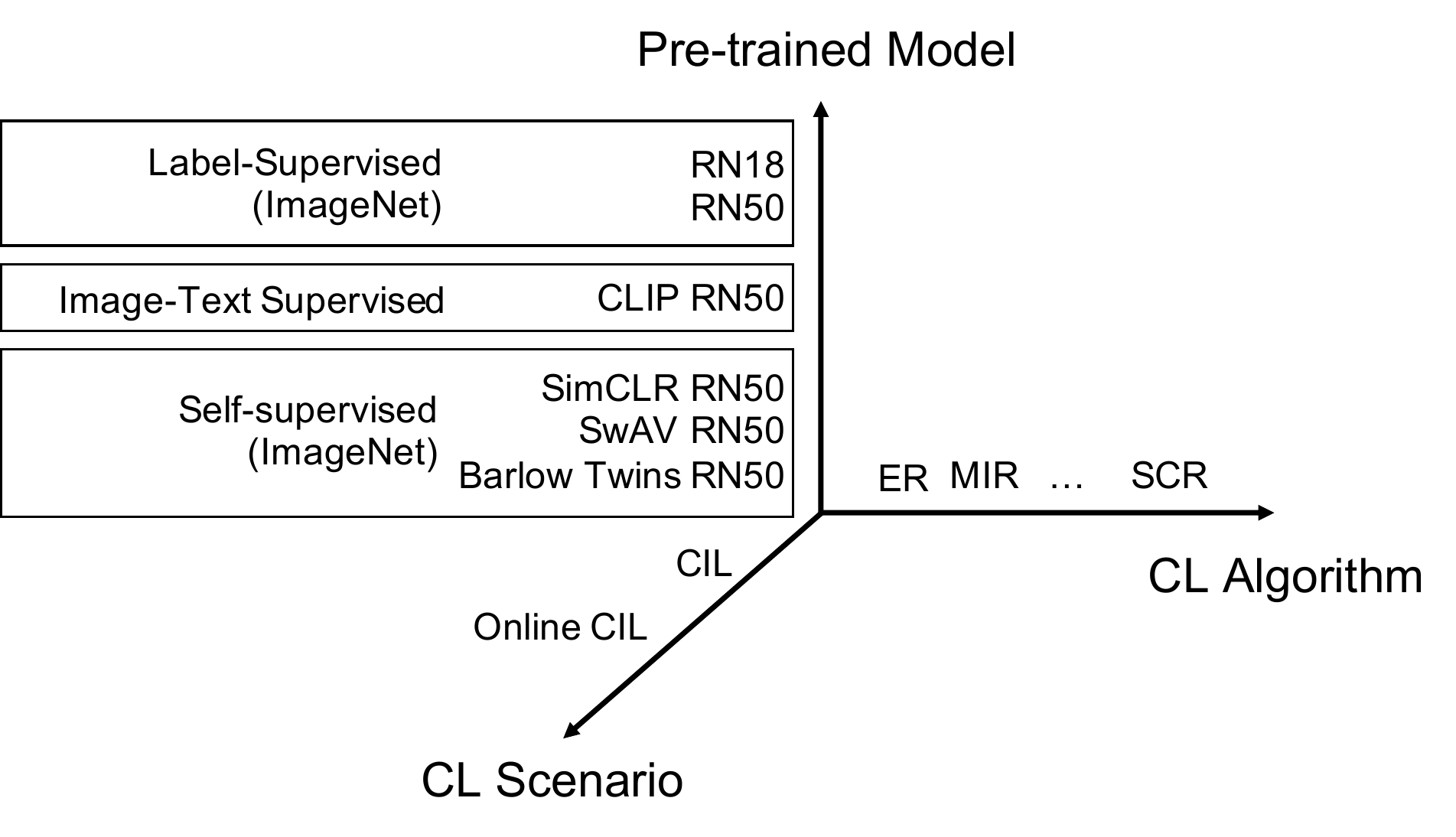}
	\scriptsize 
	\addtolength\tabcolsep{2.3pt}
	\begin{tabular}{r|c}
		\toprule
		Axis & Configurations \\
		\midrule
		Pre-trained Models (7) & Reduced RN18, RN18, RN50, CLIP RN50, \\
		& SimCLR RN50, SwAV RN50, Barlow Twins RN50\\
		\midrule
		CL Algorithms (11) & {ER, MIR, GSS, iCaRL, GDumb, SCR,} \\
		& LwF, EWC++, AGEM, Co\textsuperscript{2}L, DER++ \\
		\midrule
		CL Scenarios (2) & CIL, Online CIL \\
		\bottomrule
	\end{tabular}
	\vspace{0.2em}
	\caption{We conduct the analyses of pre-trained models in CL by dissecting the space into three axes: 1) different pre-trained models, 2) different CL algorithms, and 3) different CL scenarios.}
	\label{tab:axes}
\end{table}
\section{Related Work}
\label{sec:related}

\noindent\textbf{Continual Learning Scenarios.}
\noindent A large portion of CL literature focuses on incremental learning, which can be further divided into three different scenarios -- task, domain, and class incremental learning~\cite{three_il}.
Amongst, the most challenging scenario is class incremental learning (CIL), where the model has to predict all previously seen classes with a single head in the absence of task information.
Most recent work~\cite{ova_inn,gem,hayes2020remind} has investigated this setting.

However, being able to iterate over the entire data of a specific task for multiple epochs is not realistic~\cite{agem, gdumb}. To this end, an online version of CIL is proposed~\cite{agem, cope}, where the model trains in an online fashion and thereby can only have access to each example once.
In this work, we also mainly investigate pre-trained models in online CIL but also report results in CIL for several representative algorithms.

\noindent\textbf{Continual Learning Methods.}
\noindent According to~\cite{de2019continual}, continual learning approaches can be divided into three classes: regularization, parameter isolation, and replay methods.
Regularization methods~\cite{ewc, si, Aljundi_2018_ECCV} prevent the learned parameters from deviating too much to prevent forgetting.
Parameter isolation methods counter forgetting completely by dedicating a non-overlapping set of parameters to each task~\cite{xu2018reinforced, rusu2016progressive}.
Replay methods either store previous instances~\cite{cndpm, icarl, cope, hayes2020remind} or generate pseudo-instances~\cite{van2020brain, cong2020gan, deep_gen_replay} on the fly for replay to alleviate forgetting.

While the aforementioned approaches all show promising results in different CL scenarios, we specifically explore regularization and memory replay-based methods, given their popularity in recent literature. And we study the behaviors of pre-trained models on these methods.

\noindent\textbf{Continual Learning with Pre-trained Models.}
While most of the CL work investigates training the learner from scratch~\cite{cope, cndpm, ewc, icarl, si, gem, gss}, there is also some work that initializes the learner from a pre-trained model~\cite{agem, cbrs, expert_gate, ova_inn, piggyback, ar1}.
They harness pre-trained models for reasons such as coping with data scarcity of the downstream task~\cite{agem} and simulating prior knowledge of the continual learner~\cite{ova_inn}.
However, they \textbf{do not} 1) systematically show the substantial benefits of pre-trained models over from-scratch trained models, 2) investigate different types of pre-trained models or fine-tuning strategies, or 3) investigate pre-trained models on different CL scenarios (incremental and online learning).
Note that we claim no contribution to be the first to apply pre-trained models on CL, but rather study the aforementioned aspects {\em comprehensively}.
\section{Methodology}
We mainly focus on online class incremental learning (CIL), which is formally defined in Sec.~\ref{subsec:formulation}.
Next, we discuss various pre-trained models and how we leverage them (Sec.~\ref{subsec:finetune}).
In Sec.~\ref{subsec:two_stage}, we introduce the two-stage training pipeline that combines online training and offline training.

\subsection{Problem Formulation}
\label{subsec:formulation}
The most widely adopted continual learning scenarios are 1) task incremental learning, 2) domain incremental learning, and 3) class incremental learning (CIL).
Amongst, CIL is the most challenging and draws the most attention, for its closer resemblance to real-world scenarios, where the model is required to make predictions on all classes seen so far with no task identifiers given (we refer interested readers to~\cite{three_il} for more details).
In this paper, we focus on a more difficult scenario -- \textit{online CIL}, where the model can only have access to the data once unless with a replay buffer. In other words, the model \textbf{can not} iterate over the data of the current task for multiple epochs, which is common in CIL. The experiments of the remaining paper are based on online CIL, unless noted otherwise (we also evaluate CIL).

Formally, we define the problem as follows.
$C_{\mathrm{total}}$ classes are split into $N$ tasks, and each task $t$ contains $C_t$ non-overlapping classes (e.g., CIFAR100 is divided into 20 tasks, with each task containing 5 unique classes).
The model is presented with $N$ tasks sequentially.
Each task is a data stream $\{S_t | 0<t<=N\}$ that presents a mini-batch of samples to the model at a time.
Each sample belongs to one of the $C_t$ unique classes, which are non-overlapping to other tasks.
We explore pre-trained models with $N=20$ tasks and present the results in Sec.~\ref{sec:diff_models}.

\subsection{Fine-Tuning Strategy} \label{subsec:finetune}
When using a pre-trained model, we initialize the model with the pre-trained weights.
Then, we fine-tune the model in either 1) supervised, or 2) self-supervised manner.

For the self-supervised fine-tuning, we experiment with the SimCLR pre-trained RN50 and fine-tune it with the SimCLR loss.
Specifically, we leverage a replay buffer to store images and labels, which are then used to train the classifier on top of the fine-tuned feature representation at the end of each CL task.
Note that we train the SimCLR feature with both images sampled from the memory and images from the data stream.

\subsection{Two-Stage Pipeline} \label{subsec:two_stage}
We combine the two-stage training pipeline proposed in~\cite{two_stage} with pre-trained models to build a strong baseline for continual learning with pre-training.

The two-stage pipeline divides the learning process into two phases -- a \textit{streaming} phase where the learner (sampler) is exposed to the data stream, and an \textit{offline} phase where the learner learns from the memory. 
Similar to another widely used method GDumb~\cite{gdumb}, the two-stage pipeline trains offline on samples in the memory after all data have been streamed. Specifically, we iterate over samples in the memory for $30$ epochs after the streaming phase.
However, GDumb performs no learning in the streaming phase and discards most of the data without learning from them, making it sub-optimal.
By contrast, the two-stage pipeline improves over GDumb, through training the model on the data while storing them at the same time.

We found that this simple two-stage pipeline is particularly effective when coupled with pre-trained models.
With the two-stage training, ER~\cite{er} could outperform the best-performing models, when all leverage a pre-trained model.
\section{Experimental Setting}
 
\subsection{Datasets}
We experiment on five datasets. Classes in each dataset are equally and randomly split into 20 tasks with no overlaps. The orderings are random, but kept the same across different experiments.

\smallskip
\noindent\textbf{Split CIFAR100.}
We follow the suggested train and test split, where each category has 500 training and 100 test data~\cite{cifar100}.

\noindent\textbf{Split CUB200.}
CUB200~\cite{cub200} contains 5594 training ($\sim$30 per category) and 6194 test images for 200 different bird species.

\noindent\textbf{Split Mini-ImageNet.}
Mini-ImageNet~\cite{mini_imagenet} contains 100 subcategories from the original 1000 categories in ImageNet~\cite{deng2009imagenet}. The 100 categories are divided into 64 training, 20 validation, and 16 test categories. We follow~\cite{scr} and re-purpose all 100 categories for training and set aside 100 images per category for testing. In total, it contains 50k and 10k images for training and testing, respectively. Note the data overlap of Split Mini-ImageNet with ImageNet which is used to pre-train ResNets -- this experiment is {\em purposefully} designed to evaluate the scenario where the dataset is extremely similar to the data used for pre-training.

\noindent\textbf{Split FGVC-Aircraft.}
FGVC-Aircraft~\cite{aircraft} contains 100 aircraft model variants, each with 100 images. We combine the suggested train and validation split for training, and the test split for testing. In total, there are 6667 training images ($\sim$66 images per category) and 3333 test images. 

\noindent\textbf{Split QuickDraw.}
QuickDraw~\cite{quickdraw} contains 50M drawings across 345 categories. We randomly select 100 categories, from which we randomly sample 100 images per category (50 for training and 50 for testing) to create a dataset of 10000 images in total.

\subsection{CL Algorithms}
We explore 11 different CL algorithms, including both regularization-based methods (in \textit{Italics}) that require no memory and replay-based methods.

\smallskip
\noindent\textbf{ER.} Experience Replay~\cite{er} stores samples from the data stream with reservoir sampling and retrieves stored samples randomly. 

\noindent\textbf{MIR.} Maximally Interfered Retrieval~\cite{maximally} performs a pseudo-update on the incoming data and retrieves samples stored in the memory whose loss increases most.

\noindent\textbf{GSS.} Gradient-based Sample Selection~\cite{gss} attempts to diversify the gradients of the samples in the memory.

\noindent\textbf{\textit{LwF}.} Learning without Forgetting~\cite{lwf} applies a knowledge distillation loss from the previous model when learning the new task. 

\noindent\textbf{iCaRL.} Incremental Classifier and Representation Learning~\cite{icarl} stores a small set of images for different classes and learns features continually with distillation and classification losses.

\noindent\textbf{\textit{EWC++}.} Elastic Weight Consolidation++~\cite{ewc++} is an online version of EWC~\cite{ewc} that constrains updates to important parameters for the previous tasks.

\noindent\textbf{GDumb.} Termed Greedy Sampler and Dumb Learner~\cite{gdumb} randomly samples data to store during the streaming phase and only performs training when queried.

\noindent\textbf{AGEM.} Averaged Gradient Episodic Memory~\cite{agem} constrains gradients with samples in the memory.

\noindent\textbf{SCR.} Supervised Contrastive Replay~\cite{scr} applies supervised contrastive loss in addition to the cross-entropy loss.

\noindent\textbf{DER++.} Dark Experience Replay++~\cite{der} matches the logits of the network throughout the past.

\noindent\textbf{Co\textsuperscript{2}L}. Contrastive Continual Learning~\cite{co2l} learns and maintains features in a self-supervised fashion.
Re-purposing Co$^2$L for \textit{online} CIL, we train the feature encoder on each task for one epoch. At the end of the streaming phase, we train the classifier using the stored buffer data for one epoch. 

\begin{table*}
    \centering
    {Split CIFAR100 \\}
    \small
    \setlength{\tabcolsep}{4.75pt}
    \begin{tabular}{cccccccccc}
        \toprule
        Model & ER\cite{er} & MIR\cite{maximally} & GSS\cite{gss} & \textit{LwF}\cite{lwf} & iCaRL\cite{icarl} & \textit{EWC++}\cite{ewc++} & GDumb\cite{gdumb} & AGEM\cite{agem} & SCR\cite{scr} \\
        \midrule
        R-RN18 & 9.07±{\scriptsize1.31} & 8.03±{\scriptsize0.78} & 6.86±{\scriptsize0.60} & 8.44±{\scriptsize0.82} & 14.26±{\scriptsize0.79} & 1.00±{\scriptsize0.00} & 9.80±{\scriptsize0.46} & 3.00±{\scriptsize0.47} & \textbf{25.80±{\scriptsize0.99}} \\
        RN18 & 43.69±{\scriptsize1.67} & 42.02±{\scriptsize1.53} & 25.59±{\scriptsize0.45} & 23.40±{\scriptsize0.12} & \textbf{56.64±{\scriptsize0.23}} & 5.36±{\scriptsize0.26} & 46.76±{\scriptsize0.73} & 4.72±{\scriptsize0.21} & 51.93±{\scriptsize0.06} \\
        \midrule
        $\Delta$ & +34.62 & +33.99 & +18.73 & +14.96 & \textbf{+42.38} & +4.36 & +36.96 & +1.72 & +26.13 \\
        \bottomrule
    \end{tabular}
    \hfill\\ \vspace{0.3cm}
    
    {Split CUB200 \\}
    \setlength{\tabcolsep}{6.5pt}
    \begin{tabular}{cccccccccc}
        \toprule
        Model & ER & MIR & GSS & \textit{LwF} & iCaRL & \textit{EWC++} & GDumb & AGEM & SCR \\
        \midrule
        R-RN18 & 1.24±{\scriptsize0.11} & 1.44±{\scriptsize0.08} & 1.46±{\scriptsize0.22} & 1.47±{\scriptsize0.11} & 1.82±{\scriptsize0.24} & 0.80±{\scriptsize0.20} & 4.49±{\scriptsize0.56} & 0.67±{\scriptsize0.12} & \textbf{5.64±{\scriptsize0.75}} \\
        RN18 & 21.05±{\scriptsize1.07} & 20.95±{\scriptsize0.66} & 17.65±{\scriptsize0.45} & 6.79±{\scriptsize0.36} & 39.95±{\scriptsize1.43} & 4.47±{\scriptsize0.10} & 38.63±{\scriptsize0.44} & 4.59±{\scriptsize0.30} & \textbf{43.03±{\scriptsize1.80}} \\
        \midrule
        $\Delta$ & +19.81 & +19.51 & +16.19 & +5.32 & \textbf{+38.13} & +3.67 & +34.14 & +3.92 & +37.39\\
        \bottomrule
    \end{tabular}
    \hfill\\ \vspace{0.3cm}
    
    {Split Mini-ImageNet \\}
    \setlength{\tabcolsep}{6.5pt}
    \begin{tabular}{cccccccccc}
        \toprule
        Model & ER & MIR & GSS & \textit{LwF} & iCaRL & \textit{EWC++} & GDumb & AGEM & SCR \\
        \midrule
        R-RN18 & 8.56±{\scriptsize0.24} & 8.00±{\scriptsize0.82} & 6.74±{\scriptsize0.15} & 7.58±{\scriptsize0.65} & 11.61±{\scriptsize0.78} & 1.00±{\scriptsize0.00} & 7.01±{\scriptsize0.40} & 3.04±{\scriptsize0.21} & \textbf{33.87±{\scriptsize1.84}} \\
        RN18 & 56.91±{\scriptsize0.54} & 54.96±{\scriptsize0.46} & 25.74±{\scriptsize4.53} & 20.41±{\scriptsize0.99} & \textbf{72.40±{\scriptsize0.52}} & 4.79±{\scriptsize0.14} & 40.00±{\scriptsize0.37} & 5.23±{\scriptsize0.41} & 67.94±{\scriptsize0.11} \\
        \midrule
        $\Delta$ & +48.35 & +46.96 & +19.00 & +12.83 & \textbf{+60.79} & +3.79 & +29.99 & +2.19 & +34.07\\
        \bottomrule
    \end{tabular}
    \vspace{0.2em}
    
    \caption{
    \textbf{Accuracy in online CIL.} Different CL algorithms benefit from a pre-trained model very differently, and the comparison results between algorithms change when they are initialized from a pre-trained model. For instance, iCaRL outperforms SCR, the best-performing model when trained from scratch, on Split CIFAR100 (56.64 vs. 51.93) and Split Mini-ImageNet (72.40 vs. 67.94).
    This indicates that training from scratch does not serve as a fairground for comparison between different algorithms, in addition to its poor applicability to complex datasets.
    R-RN18 and RN18 stand for Reduced ResNet18 trained from scratch and ImageNet pre-trained ResNet18, respectively.
    }
    \label{tab:cifar100}
\end{table*}

\begin{table*}
    \centering
    \footnotesize
    \setlength{\tabcolsep}{2pt}
    \begin{tabular}{cccccccccc|cc}
        \toprule
        Model & ER\cite{er} & MIR\cite{maximally} & GSS\cite{gss} & \textit{LwF}~\cite{lwf} & iCaRL\cite{icarl} & \textit{EWC++}\cite{ewc++} & GDumb\cite{gdumb} & AGEM\cite{agem} & SCR\cite{scr} & DER++\cite{der} & Co\textsuperscript{2}L\cite{co2l} \\
        \midrule
        R-RN18 & 9.07±{\scriptsize 1.31} & 8.03±{\scriptsize0.78} & 6.86±{\scriptsize0.60} & 8.44±{\scriptsize0.82} & 14.26±{\scriptsize0.79} & 1.00±{\scriptsize0.00} & 9.80±{\scriptsize0.46} & 3.00±{\scriptsize0.47} & \textbf{25.80±{\scriptsize0.99}} & 15.72±{\scriptsize1.33} & 2.31±{\scriptsize0.64} \\
        
        RN18 & 43.69±{\scriptsize1.67} & 42.02±{\scriptsize1.53} & 25.59±{\scriptsize0.45} & 23.40±{\scriptsize0.12} & \textbf{56.64±{\scriptsize0.23}} & 5.36±{\scriptsize0.26} & 46.76±{\scriptsize0.73} & 4.72±{\scriptsize0.21} & 51.93±{\scriptsize0.06} & 44.42±{\scriptsize1.29} & 
        5.68±{\scriptsize3.19} \\
        
        RN50 & 50.88±{\scriptsize0.84} & 50.20±{\scriptsize2.80} & 31.53±{\scriptsize3.37} & 26.68±{\scriptsize0.97} & \textbf{59.20±{\scriptsize0.33}} & 3.47±{\scriptsize1.42} & 57.37±{\scriptsize0.21} & 4.49±{\scriptsize0.27} & 56.22±{\scriptsize0.42} & 49.37±{\scriptsize1.36} & 8.57±{\scriptsize0.57} \\
        \midrule
        
        CLIP & 52.31±{\scriptsize2.66} & \textbf{55.38±{\scriptsize0.83}} & 25.60±{\scriptsize4.50} & 37.21±{\scriptsize2.14} & \red{26.05±{\scriptsize12.33}} & ---\textsuperscript{*} & 55.10±{\scriptsize0.22} & 17.22±{\scriptsize2.52} & \red{30.93±{\scriptsize5.44}} & \textbf{53.01±{\scriptsize0.18}} & 1.12±{\scriptsize0.16} \\
        \midrule
        
        SimCLR RN50 & 37.04±{\scriptsize 0.48} & 40.01±{\scriptsize 1.86} & 16.32±{\scriptsize 1.52} & \red{3.40±{\scriptsize 0.17}} & 33.76±{\scriptsize 0.84} & 6.39±{\scriptsize 0.82} & 24.63±{\scriptsize 0.84} & 3.87±{\scriptsize 0.32} & \textbf{52.60±{\scriptsize 0.22}} & 15.63±{\scriptsize0.96} & 
        1.44±{\scriptsize0.45} \\
        
        SwAV RN50 & 38.32±{\scriptsize 0.11} & 40.97±{\scriptsize 0.36} & 15.00±{\scriptsize 0.30} & \red{3.32±{\scriptsize 0.45}} & 24.29±{\scriptsize 1.32} & 3.58±{\scriptsize 3.00} & 20.95±{\scriptsize 1.33} & 3.86±{\scriptsize 0.29} & \textbf{50.59±{\scriptsize 0.09}} & 20.10±{\scriptsize0.88} & 
        1.18±{\scriptsize0.26} \\
        
        B.T. RN50 & 26.15±{\scriptsize 0.62} & 18.18±{\scriptsize 1.60} & 8.38±{\scriptsize 0.23} & \red{3.70±{\scriptsize 0.16}} & 40.77±{\scriptsize 0.92} & 6.65±{\scriptsize 1.06} & 31.56±{\scriptsize 2.01} & 3.95±{\scriptsize 0.31} & \textbf{48.35±{\scriptsize 0.73}} & 5.26±{\scriptsize0.17} & 
        1.10±{\scriptsize0.10}\\
        \bottomrule
    \end{tabular}
    {\hfill\\ \scriptsize \textsuperscript{*}EWC++ fails to train with losses of \textit{nan}.}
    
    \caption{
    \textbf{Accuracy of different pre-trained models} {\em when fine-tuned in a supervised manner} on Split CIFAR100 in online CIL.
    In most cases, RN pre-trained on ImageNet (RN50 vs. CLIP RN50) in a supervised fashion (RN50 vs. SimCLR, SwAV, and Barlow Twins RN50) brings the largest accuracy increase.
    \red{Red} numbers mark pre-trained accuracy that is within/below one std. of the from-scratch counterpart, which indicates potential negative impacts.
    \textbf{Bold} numbers indicate the best accuracy amongst all methods with a specific model (e.g., 25.80 of SCR is the best within R-RN18).
    R-RN18, RN18, RN50, and CLIP stand for Reduced ResNet18 trained from scratch, ImageNet pre-trained ResNet18 and ResNet50, and CLIP pre-trained ResNet50, respectively. B.T. stands for Barlow Twins.
    \textbf{[Best viewed in color.]}}
    \label{tab:accu_diff_model}
    \vspace{-0.5em}
\end{table*}

\subsection{Pre-trained Models}
\noindent\textbf{Reduced RN18 (R-RN18).} ResNet18 whose number of channels is reduced~\cite{gem} compared with a standard one, which is used in the experiment of training from scratch.

\noindent\textbf{ImageNet Pre-trained RN18, RN50.} ResNets pre-trained on ImageNet~\cite{deng2009imagenet}.

\noindent\textbf{CLIP Pre-trained RN50.} ResNet50 pre-trained on the WebImageText dataset based on Contrastive Language–Image Pre-training (CLIP)~\cite{radford2021learning}.

\noindent\textbf{SimCLR RN50.} ResNet50 pre-trained on ImageNet with the SimCLR loss that brings closer features of different augmentations from the same image, while pushing apart those from different images~\cite{simclr}.

\noindent\textbf{SwAV RN50.} ResNet50 pre-trained on ImageNet with the SwAV mechanism that predicts the cluster assignment of a view from the representation of another one~\cite{swav}.

\noindent\textbf{Barlow Twins RN50.} ResNet50 pre-trained on ImageNet with the Barlow Twins loss that encourages the correlation of two views from the same image to be one, while discouraging that of views from different images to be zero~\cite{barlow_twins}.

\subsection{Implementation Details}
\noindent\textbf{Image Pre-processing and Architectures.}
We use the pre-trained ResNets provided in PyTorch and the CLIP Github repository\footnote{\tiny 
\url{https://github.com/openai/CLIP}}.
For SimCLR, SwAV, and Barlow Twins, we utilize pre-trained weights provided in Pytorch Lightning Bolts\footnote{\tiny \url{https://pytorch-lightning-bolts.readthedocs.io/en/stable}}.
For different CL algorithms evaluated, we adopt the publicly available code from~\cite{scr}\footnote{\tiny \url{https://github.com/RaptorMai/online-continual-learning}}.
The memory capacity for all replay-based methods is set to 1000. 

When training from scratch, we do not perform any resizing and pre-processing for most of the datasets; for Split CUB200, we resize the short side of the images to 224, with a random crop for training and a center crop for testing.
When initializing the continual learner from a pre-trained model, we resize the short side of the image to 224, followed by a random crop and a center crop for training and testing, respectively.
Note that except random cropping, we do not apply other data augmentation.
We normalize the images by the statistics computed during pre-training on either ImageNet (RN18, RN50, SimCLR RN50, SwAV RN50, and Barlow Twins RN50) or CLIP data.

When training from scratch, we follow~\cite{scr} and use a reduced ResNet18 (R-RN18) with fewer channels and a smaller first convolution kernel.

\smallskip
\noindent\textbf{Classifier Initialization.}
For all models except CLIP RN50, we randomly initialize the classifier and set its output dimension to the number of classes in the target dataset.
For CLIP RN50, we utilize its text encoder and follow the zero-shot classification guideline provided in the CLIP Github repository to generate the classifier.
Specifically, we use the text encoder to encode text in the format of `A photo of XX,' with XX being one of the categories.
The weights of the classifier are then initialized from these $N$ text features, with $N$ being the number of classes~\cite{radford2021learning}.

\smallskip
\noindent{\textbf{Hyper-parameter Selection.}}
For the two-stage training, we do not perform an extensive hyper-parameter search on CL scenarios, but instead directly use hyper-parameters that work reasonably in offline training.
Specifically, for supervised fine-tuning, we set learning rate to 0.1, 0.01, and 1e-6 for from-scratch trained Reduced ResNet18, pre-trained ResNets (RN18, RN50, SimCLR RN50, SwAV RN50, and Barlow Twins RN50), and CLIP RN50, respectively.
For self-supervised fine-tuning with the SimCLR pre-trained RN50, we set the learning rate to 0.375.
For CIL, we train the model for 5 epochs on each task.

For all methods, the batch size is set as 10. However, when fine-tuning SimCLR in a self-supervised fashion, we use a batch size of 64 (due to GPU memory constraints).
All models except SimCLR are optimized via stochastic gradient descent (SGD) with no momentum and weight decay. For SimCLR, we follow~\cite{simclr} and apply layer-wise adaptive rate scaling (LARS)~\cite{lars} on top of the plain SGD. A momentum of $0.9$ and a weight decay of 1e-6 are applied.

For all results, we take the mean and standard deviation across three runs with different random seeds.

\section{Experimental Result}
We perform investigation on pre-trained models along three axes: 1) different CL algorithms, 2) different pre-trained models, and 3) different CL scenarios.

In Sec.~\ref{sec:diff_method}, along the \textit{CL algorithm} axis, we show varied impact of pre-trained models on different algorithms.
In Sec.~\ref{sec:diff_models}, along the \textit{pre-trained model} axis, we examine difference between pre-trained models.
In Sec.~\ref{sec:diff_scenarios}, along the \textit{CL scenario} axis, we explore different CL scenarios.
Finally, in Sec.~\ref{sec:strong_baseline}, we integrate several observations and demonstrate the proposed strong baseline.

\subsection{Comparison of Different CL Algorithms} \label{sec:diff_method}

\noindent\textbf{Pre-trained models are essential for more complex CL datasets.} \label{sec:existing}
From Table~\ref{tab:cifar100}, one can observe that while many of the CL algorithms work fine on Split CIFAR100, they witness a severe performance drop when applied to Split CUB200, a more complex dataset.
This clearly illustrates the huge benefit and the necessity of pre-trained models, if one were to deal with real-world CL problems that are likely even more difficult than Split CUB200.
In fact, we would almost always consider initializing the learner from a pre-trained model in practice, reasonably assuming that the learner has some prior knowledge~\cite{ova_inn} to achieve better performance.

\smallskip
\noindent\textbf{Varied benefits of pre-trained models on distinct CL algorithms.} \label{sec:benefit}
From Fig.~\ref{fig:teaser}, Table~\ref{tab:cifar100}, and Table~\ref{tab:accu_diff_model}, clearly, the benefits of pre-trained models are not homogeneous across different CL algorithms.
For example, while ER enjoys a 34.62\% accuracy gain, LwF only has a 14.96\% improvement, despite the similar accuracy when applied on a from-scratch trained model.
Further if we sort the performance gains brought by ImageNet RN18 on Split CIFAR100 to different algorithms, one could observe that methods enjoying the most gains are ones that do not apply too much regularization (e.g., to gradients) during training.
One such comparison is between ER and EWC++: ER exerts no additional loss other than the cross-entropy loss, while EWC++ applies a regularization loss based on the importance of network parameters, which is approximated by the diagonal of the Fisher information matrix.

\smallskip
\noindent\textbf{Superiority inconsistency between from-scratch training and pre-training.}
The unequal gains also affect comparisons between different algorithms. For instance in Table~\ref{tab:cifar100}, \textit{interestingly}, while iCaRL only achieves 14.26\% in accuracy, the comparison result with SCR (25.80\%) is overturned when both are applied on ImageNet RN18 (56.64\% of iCaRL vs. 51.93\% of SCR).

\smallskip
\noindent \textbf{Remark.} Given the observations that pre-trained models should be applied in a more complex dataset and that the superiority of different algorithms does not hold when applied to a pre-trained model, future development and evaluation of CL algorithms should take pre-trained models into consideration.

\begin{table*}
    \centering
    \small
    \setlength{\tabcolsep}{3.8pt}
    \begin{tabular}{cccccccccc}
        \toprule
        Model & ER\cite{er} & MIR\cite{maximally} & GSS\cite{gss} & \textit{LwF}\cite{lwf} & iCaRL\cite{icarl} & \textit{EWC++}\cite{ewc++} & GDumb\cite{gdumb} & AGEM\cite{agem} & SCR\cite{scr} \\
        \midrule
        R-RN18 & 52.10±{\scriptsize2.22} & 50.51±{\scriptsize2.34} & 50.69±{\scriptsize0.23} & 8.83±{\scriptsize1.31} & 12.27±{\scriptsize0.34} & \textbf{2.19±{\scriptsize0.28}} & 13.59±{\scriptsize0.14} & \textbf{59.23±{\scriptsize1.47}} & 21.00±{\scriptsize0.79} \\
        
        RN18 & 50.22±{\scriptsize1.82} & 51.52±{\scriptsize1.77} & 66.85±{\scriptsize0.98} & -2.14±{\scriptsize2.32} & 15.77±{\scriptsize0.40} & 89.37±{\scriptsize0.58} & 19.56±{\scriptsize0.78} & 90.13±{\scriptsize0.50} & 22.84±{\scriptsize0.85} \\
        
        RN50 & 42.93±{\scriptsize0.67} & 40.92±{\scriptsize2.44} & 58.88±{\scriptsize3.40} & \textbf{-3.42±{\scriptsize1.18}} & 14.65±{\scriptsize0.65} & 87.36±{\scriptsize2.76} & 17.02±{\scriptsize0.48} & 90.26±{\scriptsize0.40} & 27.52±{\scriptsize0.20} \\
        \midrule
        
        CLIP & \textbf{35.51±{\scriptsize2.90}} & \textbf{30.22±{\scriptsize1.98}} & \textbf{10.53±{\scriptsize1.95}} & -1.81±{\scriptsize0.93} & \textbf{9.73±{\scriptsize0.17}} & ---\textsuperscript{*} & \textbf{5.84±{\scriptsize0.12}} & 73.82±{\scriptsize2.53} & 15.79±{\scriptsize1.48} \\
        \midrule
        
        SimCLR RN50 & 46.88±{\scriptsize0.90} & 40.12±{\scriptsize1.80} & 59.12±{\scriptsize2.47} & -0.01±{\scriptsize0.04} & 14.88±{\scriptsize1.11} & 64.16±{\scriptsize0.23} & 22.30±{\scriptsize1.33} & 75.99±{\scriptsize1.09} & \textbf{11.14±{\scriptsize0.41}} \\
        
        SwAV RN50 & 45.85±{\scriptsize0.15} & 39.26±{\scriptsize0.39} & 60.58±{\scriptsize1.66} & -0.22±{\scriptsize0.13} & 13.39±{\scriptsize0.72} & 48.83±{\scriptsize14.62} & 24.55±{\scriptsize1.26} & 74.66±{\scriptsize1.59} & 12.20±{\scriptsize0.48} \\
        
        B.T. RN50 & 60.57±{\scriptsize0.55} & 56.80±{\scriptsize2.05} & 63.71±{\scriptsize1.38} & 0.25±{\scriptsize0.24} & 16.15±{\scriptsize0.42} & 74.19±{\scriptsize0.40} & 18.13±{\scriptsize2.20} & 78.31±{\scriptsize1.40} & 14.50±{\scriptsize0.63} \\
        \bottomrule
    \end{tabular}
    {\hfill\\ \scriptsize \textsuperscript{*}EWC++ fails to train with losses of \textit{nan}.}
    
    \caption{
    \textbf{Forgetting of different pre-trained models} {\em when fine-tuned in a supervised manner} on Split CIFAR100 in online CIL.
    \textbf{Bold} numbers indicate the least forgetting amongst all the models (column-wise comparison).
    CLIP shows consistently less forgetting compared with other ImageNet pre-trained ResNets.
    R-RN18, RN18, RN50, and CLIP stand for Reduced ResNet18 trained from scratch, ImageNet pre-trained ResNet18 and ResNet50, and CLIP pre-trained ResNet50, respectively. B.T. stands for Barlow Twins.
    }
    \label{tab:forgetting_diff_model}
\end{table*}

\begin{table*}
    \centering
    \footnotesize
    \setlength{\tabcolsep}{2.3pt}
    \begin{tabular}{c|c|ccc|c}
        \multicolumn{6}{c}{(a) {\small Experience Replay (ER)}} \\  
        \toprule
        Fine-Tuning & From-scratch & \multicolumn{3}{c|}{Supervised} & Self-supervised \\
        \midrule
        & R-RN18 & RN18 & RN50 & CLIP & SimCLR RN50 \\
        \midrule
        \multirow{2}{*}{CIL} & 8.17±1.06 & 36.21±1.17 & 44.18±2.55 & 55.44±1.34 & 34.72±4.04 \\
        & {\scriptsize(63.88±1.07)} & {\scriptsize(59.05±0.82)} & {\scriptsize(50.96±2.29)} & {\scriptsize(36.36±0.98)} & {\scriptsize(20.26±1.87)}\\ 
        \midrule
        \multirow{2}{*}{Online CIL} & 9.07±1.31 & 43.69±1.67 & 50.88±0.84 & 52.79±1.91 & 33.39±0.42\\
        & {\scriptsize(52.10±2.22)} & {\scriptsize(50.22±1.82)} & {\scriptsize(42.93±0.67)} & {\scriptsize(35.51±2.90)} & {\scriptsize(20.28±0.75)}\\ 
        \bottomrule
    \end{tabular}
    \quad
    \begin{tabular}{c|c|ccc}
        \multicolumn{5}{c}{(b) {\small Learning without Forgetting (LwF)}} \\ 
        \toprule
        Fine-Tuning & From-scratch & \multicolumn{3}{c}{Supervised} \\
        \midrule
        & R-RN18 & RN18 & RN50 & CLIP  \\
        \midrule
        \multirow{2}{*}{CIL} & 13.05±0.65 & 19.18±0.86 & 17.82±1.83 & 35.52±1.90 \\
        & {\scriptsize(8.33±4.35)} & {\scriptsize(-4.40±1.44)} & {\scriptsize(-3.50±2.10)} & {\scriptsize(-3.97±1.25)} \\ 

        \midrule
        \multirow{2}{*}{Online CIL} & 8.44±0.82 & 23.40±0.12 & 25.98±1.34 & 37.73±1.19 \\
        & {\scriptsize(8.83±1.31)} & {\scriptsize(-2.14±2.32)} & {\scriptsize(-3.42±1.18)} & {\scriptsize(-1.81±0.93)} \\ 
        \bottomrule
    \end{tabular}
    
    \vspace{0.2em}
    \caption{\textbf{Accuracy (Forgetting) of different models} on Split CIFAR100.
    (a) {\em Self-supervised fine-tuning} (SimCLR) demonstrates a lower forgetting compared with supervised fine-tuning (20.26 vs. 50.96 of RN50 in CIL).
    (a)(b) CLIP, pre-trained with image-text pairs, shows less forgetting compared with ResNets pre-trained with curated ImageNet labels.
    Numbers outside/inside parentheses are accuracy/forgetting, respectively.
    R-RN18 and RN18 stand for Reduced ResNet18 and ImageNet pre-trained ResNet18, respectively.
    }
    \label{tab:er}
\end{table*}

\subsection{Comparison of Different Pre-trained Models} \label{sec:diff_models}

\begin{figure}
    \centering
    \includegraphics[width=0.9\linewidth]{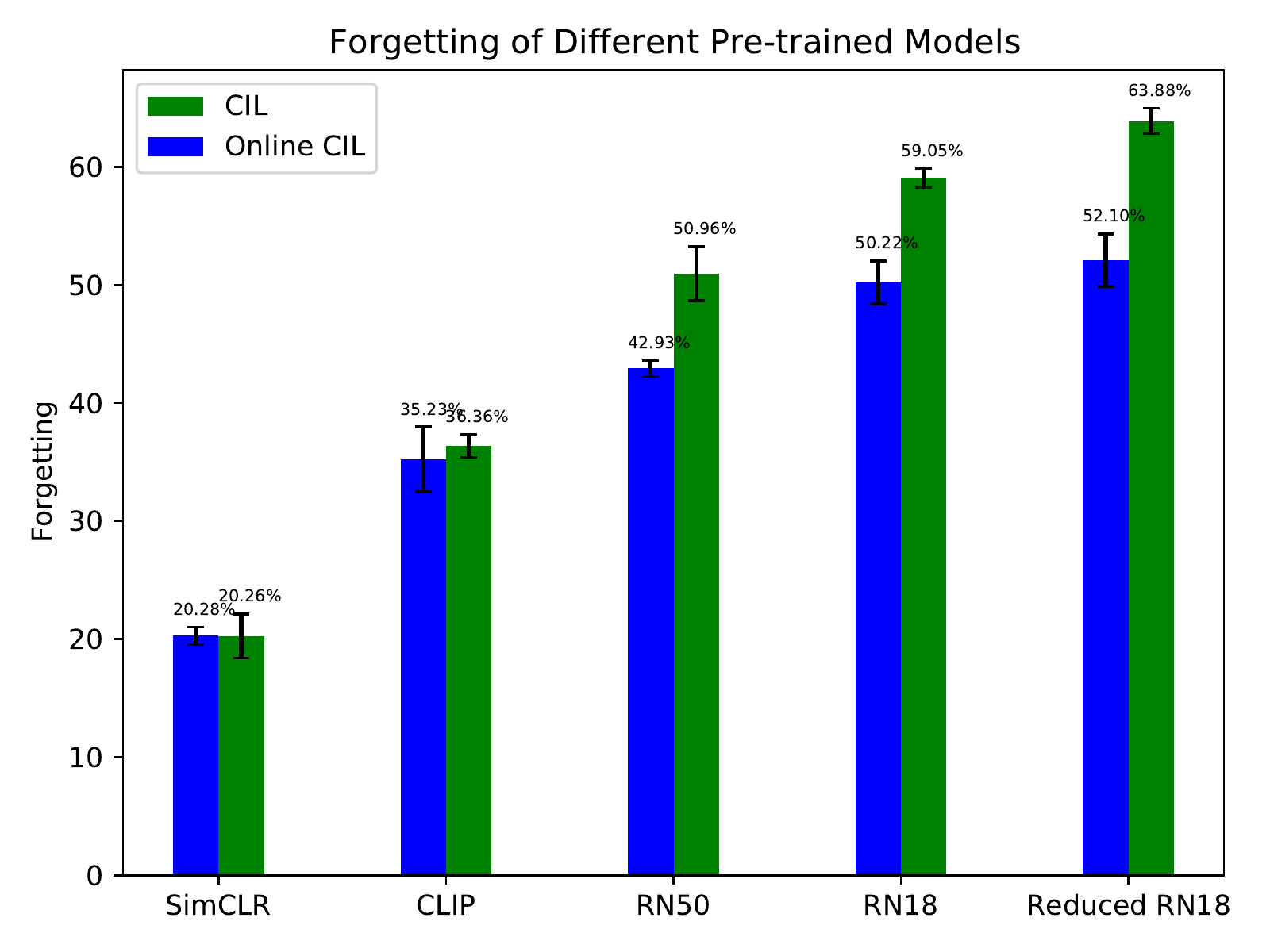}
    \caption{\textbf{Forgetting of different models.} Fine-tuning a pre-trained model in a self-supervised manner (with the SimCLR loss) largely decreases forgetting, compared with supervised fine-tuning.
    \textbf{[Best viewed in color.]}} 
    \label{fig:forgetting}
    \vspace{-1em}
\end{figure}

\noindent\textbf{Comparison on Forgetting.} \label{sec:ssl_forgetting}
In Table~\ref{tab:forgetting_diff_model}, one could observe that CLIP RN50 enjoys consistently lower forgetting, compared with training from scratch (R-RN18).
Yet, the other pre-trained models (also fine-tuned in a supervised manner) do not have the same advantage, regardless of being pre-trained in a supervised or self-supervised manner.

Table~\ref{tab:er} and Fig.~\ref{fig:forgetting} show that self-supervised fine-tuning (SimCLR) has less forgetting than supervised fine-tuning.
We identify two differences between self-supervised and supervised fine-tuning -- 1) decoupled training of the feature and classifier, and 2) self-supervision.
Our in-depth experiment in Sec.~\ref{sec:sup_forgetting} of the Supplementary Material shows that the decoupled training likely helps mitigate forgetting.

\begin{table}
    \footnotesize
    \centering
    \begin{tabular}{ccccc}
        \toprule
             & S-CIFAR100 & S-CUB200 & S-QuickDraw & S-Aircraft \\
        \midrule
        RN50 & 57.68±0.23 & \textbf{48.05±0.87} & \textbf{55.19±0.41} & \textbf{34.23±0.98}\\
        CLIP & \textbf{59.22±0.33} & 44.98±0.31 & 34.88±0.49 & 19.17±0.14\\
        \bottomrule
    \end{tabular}
    \vspace{0.2em}
    
    \caption{\textbf{ImageNet RN50 vs. CLIP RN50} in online CIL. Despite the impressive zero-shot capability, CLIP underperforms ImageNet pre-trained RN50 in three out of the four datasets, suggesting that the better pre-trained model to consider is probably still ImageNet RN. S- stands for Split. RN50 and CLIP are ImageNet ResNet50 and CLIP ResNet50, respectively.
    }
    \label{tab:clip_vs_rn50}
\end{table}

\begin{figure}
    \centering
    \includegraphics[keepaspectratio=true,width=\linewidth]{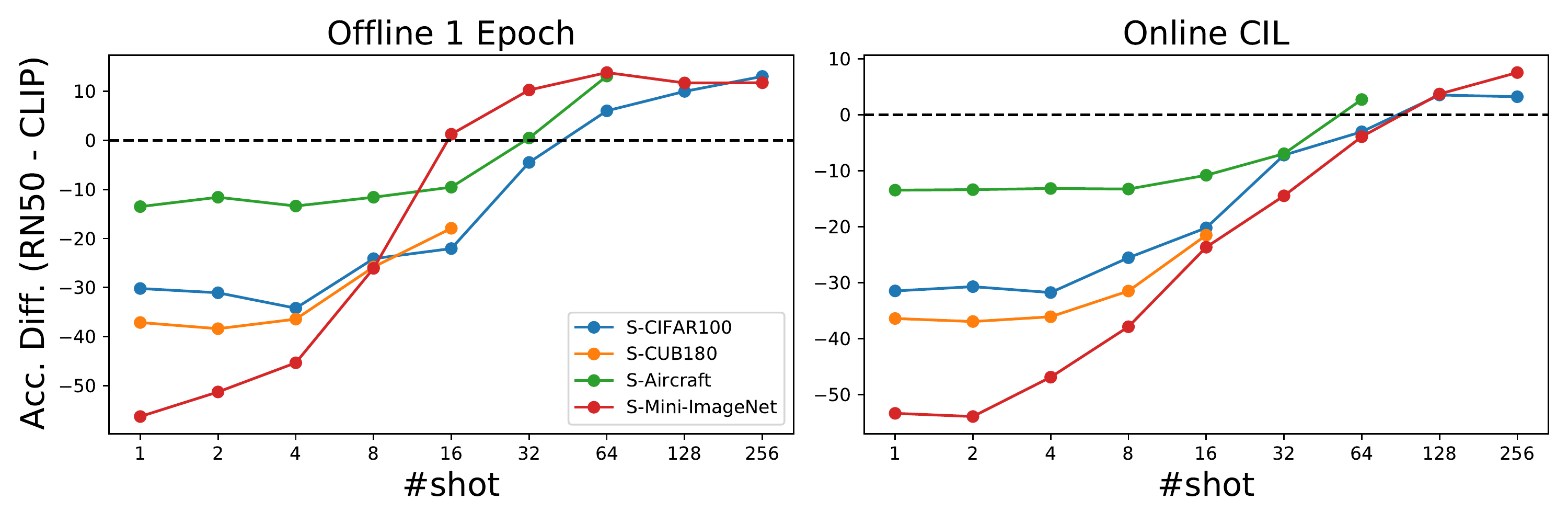}
    \caption{
    \textbf{Comparison between CLIP R50 and ImageNet RN50.} CLIP is better in the few-shot regime, while ImageNet RN50 prevails with plenty of data, as is the case for the CL scenarios considered in the paper.
    This explains the better performance when a CL learner is initialized from an ImageNet RN50 (cf. Table~\ref{tab:clip_vs_rn50}). S- stands for Split.
    \textbf{[Best viewed in color.]}
    } 
    \label{fig:clip_vs_rn50}
    \vspace{-1em}
\end{figure}

\smallskip
\noindent\textbf{CLIP vs. ImageNet Pre-trained RN50.} \label{sec:clip_vs_rn50}
From Table~\ref{tab:clip_vs_rn50}, surprisingly, despite its impressive zero-shot capability, CLIP RN50 is still worse than ImageNet RN50 in most cases. 
Based on Fig.~\ref{fig:clip_vs_rn50}, we conjecture that CLIP RN50 is better in few-shot regimes, while ImageNet RN50 prevails with plenty of data, which coincides with the CL scenarios considered in this work.

\subsection{Comparison of Different CL Scenarios} \label{sec:diff_scenarios}

From Table~\ref{tab:er}, one can observe that the performance of CIL is lower than online CIL.
While seemingly counter-intuitive, this is likely because, in CIL, the model overfits too much to the current task, which worsens the forgetting. The same behavior is also observed in~\cite{fluid}.

\subsection{A Strong Baseline for CL with Pre-training} \label{sec:strong_baseline}
We have observed that 1) algorithms that apply less regularization tend to benefit more from pre-trained models, and 2) ResNets trained on ImageNet provide a better improvement on CL in general.
Based on these insights, we propose a strong baseline for CL with pre-training. We combine the simplest ER that exerts no regularization during training, ImageNet RN50, and the two-stage training pipeline discussed in Sec.~\ref{subsec:two_stage}.

\begin{figure}
    \centering
\includegraphics[width=0.9\linewidth]{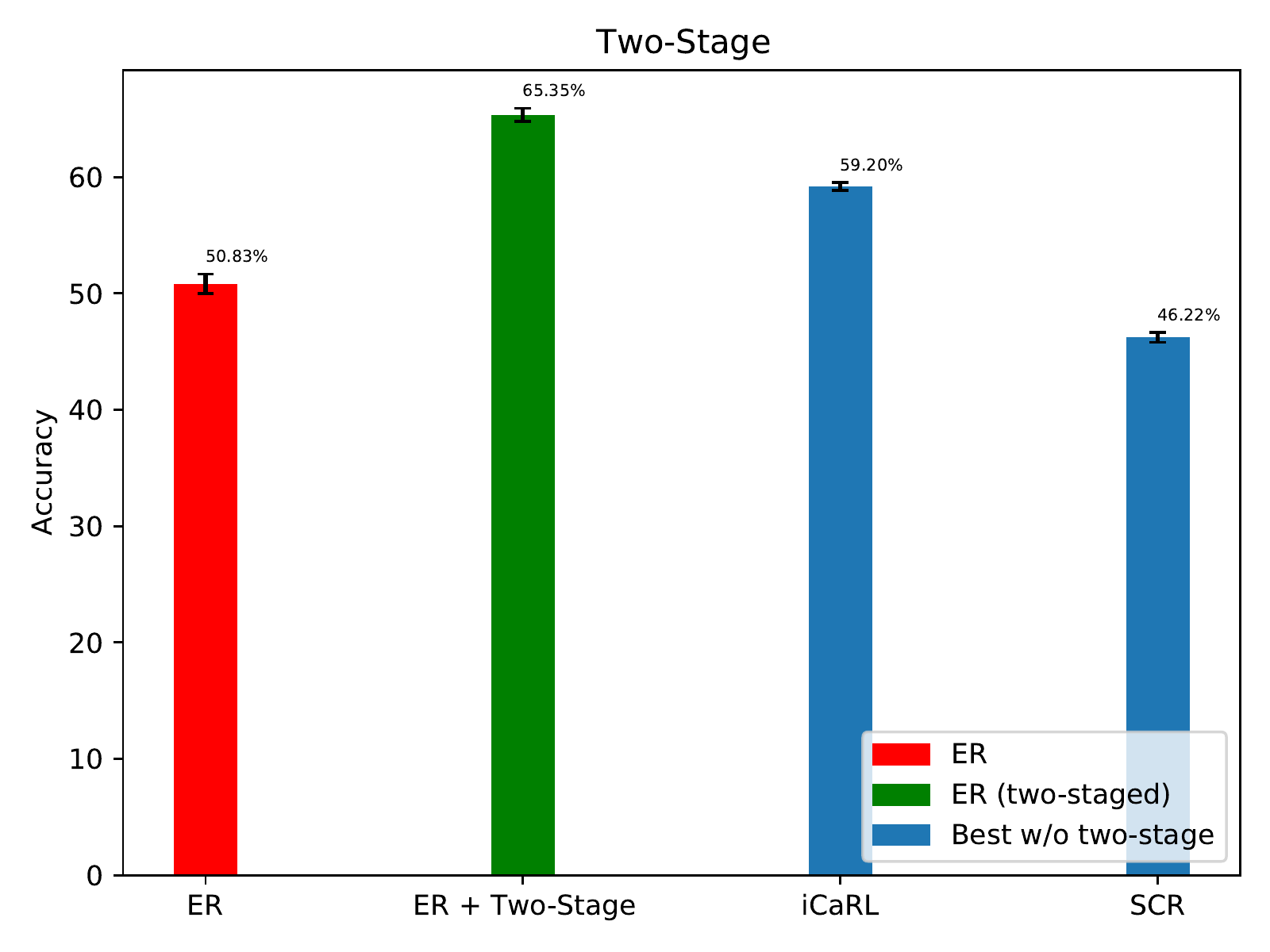}
    \caption{A simple second-staged offline training on memory data coupled with an ImageNet pre-trained ResNet50 turns a simple baseline into state of the art, suggesting the effectiveness of the proposed baseline.
    Note that SCR and iCaRL are the two best-performing methods when applied on the ImageNet pre-trained ResNet50.
    \textbf{[Best viewed in color.]}} 
    \label{fig:two_stage}
    \vspace{-1em}
\end{figure}

In Fig.~\ref{fig:two_stage} and Table~\ref{tab:two_stage}, we compare this baseline with best-performing methods, iCaRL and SCR.
We observe that with the additional offline training, the accuracy of ER increases from 50.83\% to 65.35\%. This demonstrates that a simple baseline ER, coupled with a pre-trained model, could turn a relatively na\"ive baseline into a strong baseline that even achieves the state-of-the-art result. Results of other methods with the two-stage pipeline are provided in Table~\ref{tab:two_stage_others} of the Supplementary Material, showing the {\em consistent} effectiveness in most cases. We hence recommend including this simple yet strong baseline for CL algorithm evaluation and comparison.

Also, we note that the observation we made here is different from the one in\cite{gdumb}.
In~\cite{gdumb}, for the from-scratch trained model, the authors found that simply fine-tuning in the offline phase without learning online achieves the best performance.
However, we found that when initialized from a pre-trained model, learning in both the online and offline phases achieves the best performance.
The discrepancy indicates potentially different training behaviors for CL between training from scratch and from a pre-trained model.

\vspace{-1em}
\section{Limitation and Discussion}
Although we have shown that pre-trained models bring a considerable amount of benefits to CL, they could limit the architecture choices.
Also, it is still unclear how one should choose the optimal pre-trained model and fine-tuning strategy given a CL task.
While we have shown that the optimality could depend on the downstream dataset and that ImageNet pre-trained ResNets are generally better, more research is required to have a better conclusion on this.

While the two-stage training pipeline seems promising, there are potential limitations which we have not addressed.
First of all, in many real-world scenarios, there will be no ending to the data stream and the learner has to endlessly learn new concepts.
This again reflects why we consider \textit{online} CIL and not CIL, as in most real-world scenarios, it is not applicable to iterate the training data multiple times.
Even if in a few scenarios where we do have the leisure to further train the model, there might be too many classes to fit into the memory. In such scenarios, it is questionable whether performing the additional fine-tuning on the limited classes in the memory is beneficial.

\setlength{\tabcolsep}{4pt}
\begin{table}
    \scriptsize
    \centering
    \begin{tabular}{cccccc}
    \hline
    & & ER & ER+2-stage & iCaRL & SCR \\
    \hline
    \multirow{3}{*}{S-CIFAR100} & RN18 & 43.69±1.67 & \textbf{58.59±0.31} & 56.64±0.23 & 51.93±0.06 \\
    & RN50 & 50.83±0.84 & \textbf{65.35±0.55} & 59.20±0.33 & 46.22±0.42 \\
    & CLIP & 52.31±2.66 & \textbf{59.42±0.80} & 26.05±12.33 & 30.93±5.44 \\
    \hline
    \hline
    \multirow{3}{*}{S-CUB200} & RN18 & 21.05±1.07 & \textbf{52.32±0.98} & 39.95±1.43 & 43.03±1.80 \\
    & RN50 & 48.05±0.87 & \textbf{59.88±0.30} & 43.69±0.79 & 50.77±0.80 \\
    & CLIP & 42.08±0.36 & \textbf{43.91±0.88} & 3.55±0.86 & 4.60±0.00 \\
    \hline
    \end{tabular}
    \vspace{0.2em}
    
    \caption{\textbf{Two-stage training pipeline.} Building on top of a pre-trained model and the two-stage training pipeline, a simple ER baseline becomes a state-of-the-art approach. This further suggests that CL algorithms should be developed and evaluated with pre-trained models taken into consideration.
    S- stands for Split.}
    \label{tab:two_stage}
    \vspace{-1em}
\end{table}
\vspace{-0.5em}
\section{Conclusion}
While most of existing continual learning algorithms have been developed in the context of from-scratch training, we show that this is not the most effective way to develop and evaluate continual learning algorithms.
As supportive observations, our extensive empirical study reveals that 1) the pre-trained models are dramatically beneficial, making them of great necessity for real-world scenarios; and 2) the benefits of a pre-trained model on different CL algorithms are vastly different. The best algorithm when trained from scratch does not necessarily perform the best when coupled with a pre-trained model.
These findings indicate that the current methodology of developing CL algorithms from scratch could be potentially less effective.

We also notice different behaviors between different pre-trained models. Somewhat surprisingly, despite being a powerful pre-trained model for its zero-shot capability, CLIP RN50 seems to underperform the supervised ImageNet RN50 in most CL cases. Further investigation into explaining different behaviors and developing model selection strategies is an interesting direction.
Another observation is that fine-tuning a SimCLR pre-trained model in a self-supervised manner has significantly lower forgetting.
The observation could be potentially leveraged to develop a hybrid method that incorporates both the self-supervised loss and the cross-entropy loss.
By doing so, one could potentially enjoy the best of both worlds, reducing forgetting of supervised fine-tuning and increasing the accuracy of self-supervised fine-tuning.

Finally, based on the observations that algorithms exerting less regularization during training benefit 
more from a pre-trained model, and that ImageNet RN50 provides more benefit than the other pre-trained models, we propose a simple yet effective baseline that achieves state-of-the-art performance on multiple datasets.

\noindent\textbf{Acknowledgement.}
{\small This work was supported in part by NSF Grant 2106825, NIFA award 2020-67021-32799, the Jump ARCHES endowment through the Health Care Engineering Systems Center, the National Center for Supercomputing Applications (NCSA) at the University of Illinois at Urbana-Champaign through the NCSA Fellows program, and the IBM-Illinois Discovery Accelerator Institute.}

{\footnotesize
\bibliographystyle{ieee_fullname}
\bibliography{egbib}
}

\renewcommand{\thefigure}{\Alph{figure}}
\renewcommand{\thetable}{\Alph{table}}
\renewcommand\thesection{\Alph{section}}

\setcounter{section}{0}
\setcounter{figure}{0}
\setcounter{table}{0}

\renewcommand{\theHtable}{Supplement.\thetable}
\renewcommand{\theHfigure}{Supplement.\thefigure}
\renewcommand{\theHsection}{Supplement.\thesection}

\newpage
\section{Additional Implementation Details}

\noindent\textbf{Detailed Setting of Each Experiment.}
All experiments in the main paper are performed on Split CIFAR100 in online CIL, if not specified otherwise. Table~\ref{tab:settings} elaborates the details of each table in the main paper.
As discussed in the Introduction, we attempt to fix one axis and analyze the effect of the other two.
For example, in Table~\ref{tab:cifar100}, we fix the CL scenario and analyze the benefits of pre-trained models. In Table~\ref{tab:er}, we fix the CL algorithm and compare pre-trained models and CL scenarios.

\smallskip
\noindent\textbf{Metrics.}
\textbf{Accuracy} refers to the all-way classification accuracy at the end of the training, e.g., 100-way classification on Split CIFAR100. 
\textbf{Forgetting} is computed by subtracting the accuracy of a task right after it is learned by the final accuracy of the same task.

\begin{table*}
    \centering
    \small
    \addtolength\tabcolsep{10pt}
	\begin{tabular}{c|c|c|c}
		\toprule
		Main paper & Models & CL Algorithm & CL Scenario \\
		\midrule
		Table~\ref{tab:cifar100} & R-RN18, RN18 & All & Online CIL \\
		Table~\ref{tab:accu_diff_model} & All & All & Online CIL \\
		Table~\ref{tab:forgetting_diff_model} & All & All but DER++, Co$^2$L & Online CIL \\
		Table~\ref{tab:er} & R-RN18, RN18, RN50, SimCLR RN50 & ER, LwF & All \\
		Table~\ref{tab:clip_vs_rn50} & RN50, CLIP & ER & Online CIL \\
		Table~\ref{tab:two_stage} & RN18, RN50, CLIP & ER, iCaRL, SCR & Online CIL \\
		\midrule
		Fig.~\ref{fig:teaser} & RN18 & All but DER++, Co$^2$L & Online CIL \\
		Fig.~\ref{fig:forgetting} & R-RN18, RN18, RN50, CLIP, SimCLR & ER & All \\
		Fig.~\ref{fig:clip_vs_rn50} & RN50, CLIP & ER & Online CIL \\
		\bottomrule
	\end{tabular}
	\caption{\textbf{Detailed configurations} of experiments in the main paper.
    We attempt to analyze pre-trained models in CL through these three axes -- different pre-trained models, different CL algorithms, and different CL scenarios.
	}
	\label{tab:settings}
\end{table*}

\section{Two-Stage Training Pipeline}
\begin{figure}
    \centering
    \includegraphics[width=\linewidth]{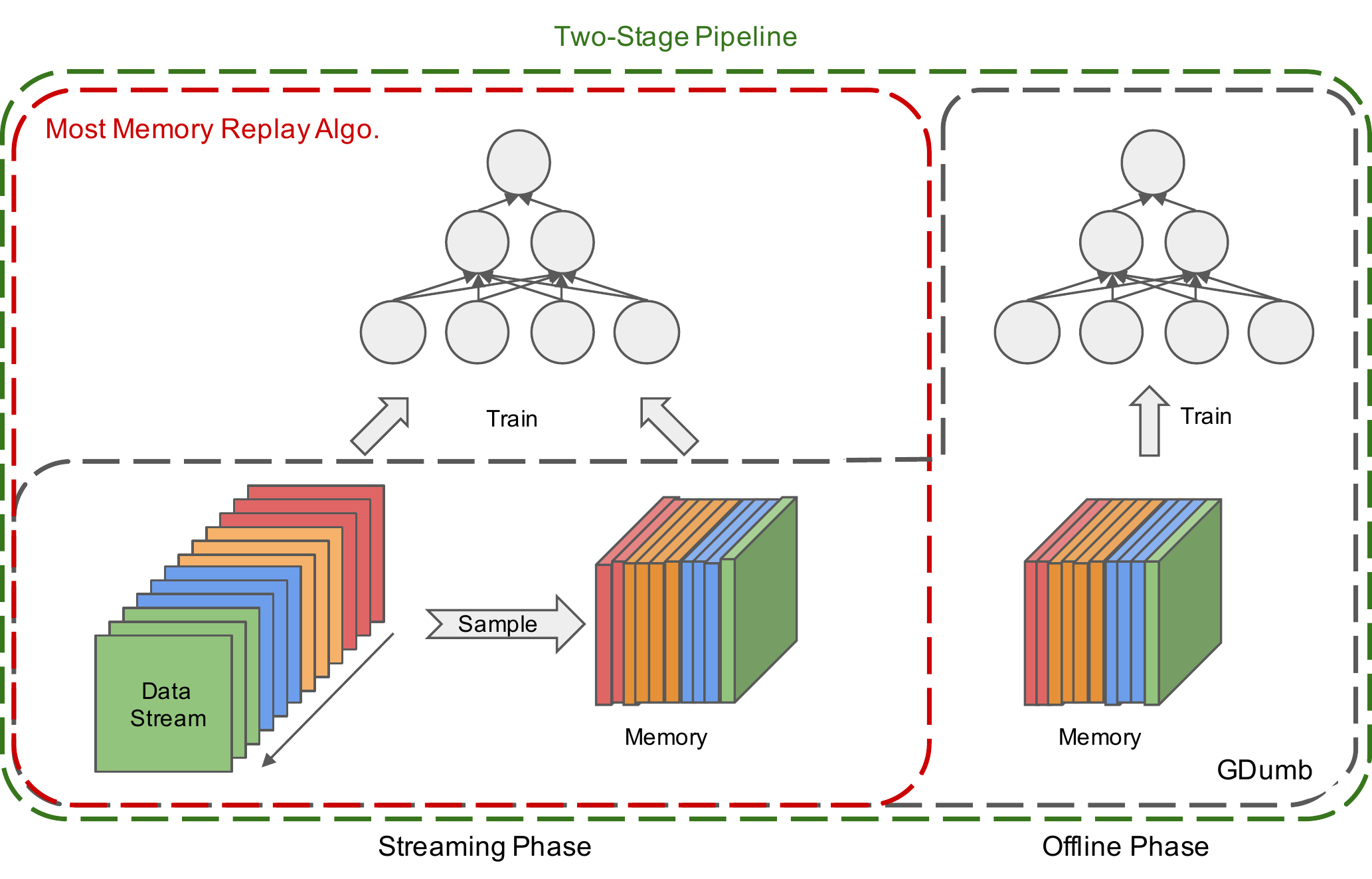}
    \caption{\textbf{Two-stage training CL pipeline.} Most memory replay methods only perform learning during the streaming phase while on the contrary, GDumb only performs learning at the end of the stream (offline phase).
    Coupled with a pre-trained model, this simple two-stage pipeline that learns in both phases converts Experience Replay (ER) into a state-of-the-art approach (cf. Table~\ref{tab:two_stage}).}
    \label{fig:framework}
\end{figure}

\noindent\textbf{Methodology.} 
The two-stage training is dependent on the underlying CL algorithm. We regard the memory as the training set and apply the exact CL algorithms on the ``memory dataset,'' but in an i.i.d. fashion for multiple (30) epochs.
See Fig.~\ref{fig:framework} for the pipeline diagram.

\smallskip
\noindent\textbf{Comprehensive Results on Different CL Algorithms.} As discussed in Table~\ref{tab:two_stage} of the main paper, the two-stage training pipeline that combines learning in the streaming phase and offline training with samples in the memory could make a simple ER method a strong baseline when coupled with a pre-trained model (ImageNet RN50).
In Table~\ref{tab:two_stage}, we further apply the two-stage training to the other CL algorithms and additionally compare the two-stage training between initialization from scratch (R-RN18+TS) and a pre-trained model (RN18+TS and RN50+TS).

\begin{table*}
    \centering
    \small
    \setlength{\tabcolsep}{5pt}
    \begin{tabular}{cccccccccc}
        \toprule
        Model & ER\cite{er} & MIR\cite{maximally} & GSS\cite{gss} & \textit{LwF}\cite{lwf} & iCaRL\cite{icarl} & \textit{EWC++}\cite{ewc++} & GDumb\cite{gdumb} & AGEM\cite{agem} & SCR\cite{scr} \\
        \midrule
        R-RN18 & 9.07±{\scriptsize1.31} & 8.03±{\scriptsize0.78} & 6.86±{\scriptsize0.60} & 8.44±{\scriptsize0.82} & 14.26±{\scriptsize0.79} & 1.00±{\scriptsize0.00} & 9.80±{\scriptsize0.46} & 3.00±{\scriptsize0.47} & \textbf{25.80±{\scriptsize0.99}} \\
        
        +TS & 14.66±{\scriptsize0.23} & 13.67±{\scriptsize0.47} & 12.67±{\scriptsize0.25} & ---\textsuperscript{*} & 14.66±{\scriptsize0.35} & ---\textsuperscript{*} & ---\textsuperscript{$\dagger$} & 12.50±{\scriptsize0.58} & \textbf{23.06±{\scriptsize0.09}} \\    
        $\Delta$ & \green{5.59} & \green{5.64} & \green{5.81} & ---\textsuperscript{*} & \green{0.40} & ---\textsuperscript{*} & ---\textsuperscript{$\dagger$} & \green{\textbf{9.50}} & \red{-2.74}\\
        \midrule
        
        RN18 & 43.69±{\scriptsize1.67} & 42.02±{\scriptsize1.53} & 25.59±{\scriptsize0.45} & 23.40±{\scriptsize0.12} & \textbf{56.64±{\scriptsize0.23}} & 5.36±{\scriptsize0.26} & 46.76±{\scriptsize0.73} & 4.72±{\scriptsize0.21} & 51.93±{\scriptsize0.06} \\
        
        +TS & 58.59±{\scriptsize0.31} & 56.64±{\scriptsize0.14} & 37.08±{\scriptsize1.45}
 & ---\textsuperscript{*} & 58.66±{\scriptsize1.07} & ---\textsuperscript{*} & ---\textsuperscript{$\dagger$} & 57.98±{\scriptsize0.57} & 49.55±{\scriptsize0.29} \\  
        $\Delta$ & \green{14.90} & \green{14.62} & \green{11.49} & ---\textsuperscript{*} & \green{2.02} & ---\textsuperscript{*} & ---\textsuperscript{$\dagger$} & \textbf{\green{53.26}} & \red{-2.38} \\
        \midrule
        
        RN50 & 50.88±{\scriptsize0.84} & 50.20±{\scriptsize2.80} & 31.53±{\scriptsize3.37} & 26.68±{\scriptsize0.97} & \textbf{59.20±{\scriptsize0.33}} & 3.47±{\scriptsize1.42} & 57.37±{\scriptsize0.21} & 4.49±{\scriptsize0.27} & 56.22±{\scriptsize0.42} \\
        
        +TS & \textbf{65.35±{\scriptsize0.55}} & 62.87±{\scriptsize0.63} & 52.03±{\scriptsize0.62}\textsuperscript{$\star$}
 & ---\textsuperscript{*} & 60.44±{\scriptsize0.13} & ---\textsuperscript{*} & ---\textsuperscript{$\dagger$} & 62.76±{\scriptsize0.54} & 51.55±{\scriptsize0.24} \\
 
        $\Delta$ & \green{14.47} & \green{12.67} & \green{20.5} & ---\textsuperscript{*} & \green{1.24} & ---\textsuperscript{*} & ---\textsuperscript{$\dagger$} & \textbf{\green{58.27}} & \red{-5.92} \\
        \midrule
        
        \bottomrule
    \end{tabular}
    \\
    {\scriptsize
    *We do not apply the two-stage training to LwF and EWC++, because no memory buffer is employed.
    \\
    $\dagger$GDumb is essentially the second stage;
    }

    \smallskip
    \caption{
    \textbf{Two-stage accuracy} on Split CIFAR100 in online CIL.
    While the two-stage training pipeline is generally beneficial, a performance drop is present in SCR (-2.38).
    ER shows the best performance despite its simplicity, when coupled with ImageNet RN50 and the two-stage training pipeline.
    The \green{green} numbers indicate a positive accuracy increase while the \red{red} numbers indicate a decrease, when the two-stage training pipeline is applied.
    \textbf{Bold} numbers indicate the best accuracy amongst all methods with a specific setting (e.g., 25.80 of SCR is the best with R-RN18).
    R-RN18 and RN18 stand for Reduced ResNet18 trained from scratch and ImageNet pre-trained ResNet18, respectively.
    \textbf{[Best viewed in color.]}
    }
    \label{tab:two_stage_others}
\end{table*}

Three observations are made here. 1) The absolute improvement brought by the two-stage training pipeline is more pronounced with an ImageNet RN (RN18+TS vs. RN18 and RN50+TS vs. RN50) compared with from-scratch training (R-RN18+TS vs. R-RN18).
2) The two-stage training pipeline is generally beneficial across CL algorithms, despite its slightly negative impact on SCR.
3) \textbf{ER shows the best performance} despite its simplicity, when coupled with ImageNet RN50 and the two-stage training pipeline.
This clearly shows the significance of our observations (minimum regularization benefits more from a pre-trained model; ImageNet RN50 generally outperforms CLIP RN50), which facilitates the proposed strong baseline. 

\section{Discussion on Forgetting} \label{sec:sup_forgetting}
\noindent\textbf{Less Forgetting with CLIP RN50.}
In Table~\ref{tab:forgetting_diff_model} of the main paper, we discussed that CLIP shows less forgetting, compared with ResNets trained with the ImageNet data in a supervised manner.
We conjecture the lower forgetting might be attributed to 1) the lower learning rate required for CLIP to train successfully, and 2) the feature normalization that projects all features on a unit hyper-sphere, potentially serving as a form of regularization.

\smallskip
\noindent\textbf{Less Forgetting with Self-supervised Fine-tuning.}
In Table~\ref{tab:er} of the main paper, we observed that self-supervised fine-tuning (with the SimCLR loss) shows less forgetting compared with supervised counterparts in the downstream task.
Here, we attempt a more in-depth examination.

Self-supervised fine-tuning of SimCLR involves 1) self-supervised update of features and 2) supervised training of the classifier.
The difference between SimCLR and fine-tuning networks in a supervised manner is two-fold: 1) the decoupling of feature and classifier training, and 2) the features are learned in a self-supervised fashion.
We attempt to isolate the effect of the decoupling mechanism. To achieve this, we train RN18 via the usual supervised cross-entropy loss with the ER method.
However, at the end of each task, we discard the learned classifier and instead train a new one from scratch with the samples in ER (as we do when fine-tuning SimCLR RN50 in a self-supervised manner).
Doing so, the forgetting is now 29.13±0.97, which is significantly lower than its counterpart ImageNet RN50 (42.93±0.67 as in Table~\ref{tab:forgetting_diff_model} of the main paper) whose feature and classifier are trained jointly.

This potentially indicates that only a minor part of the forgetting results from the supervised training of the features and it is the joint training of the features and the classifier that causes the most forgetting.
Such an observation might be loosely connected to~\cite{longtail}, where it was found harmless to train features with long-tailed distributed data. However, when training the classifier, a re-balanced dataset is instead utilized.
This is, to some degree, analogous to CL with a replay buffer since during the streaming phase, most data come from the current task (set of classes) while previous classes only account for a small portion (fairly assuming the replay is relatively small compared with the size of data from the current task).
In other words, similarly, during the streaming phase in CL, one conducts feature learning with a quasi-long-tailed distribution.

\smallskip
\noindent\textbf{Connection to Two-Stage Pipeline.}
The aforementioned experiment potentially explains why the two-stage mechanism further improves the performance.
Since training the classifier jointly during the streaming phase shows the most forgetting, the second (offline) stage mitigates the forgetting by learning the classifier with a balanced sample set (samples in the memory). 
This could also be analogous to the second step of~\cite{longtail}, training the classifier with a balanced sample set.

\begin{table}
    \centering
    \small
    \setlength{\tabcolsep}{10pt}
    \begin{tabular}{c|cc}
        \toprule
        Model  & Co\textsuperscript{2}L~\cite{ co2l} & Co\textsuperscript{2}L + Two-Stage \\
        \midrule
        R-RN18 & 2.31±{\scriptsize0.64} & 6.30±{\scriptsize0.46} \\
        
        RN18 & 5.68±{\scriptsize3.19} & 38.71±{\scriptsize1.05}\\
        
        RN50 & 8.57±{\scriptsize0.57} & 43.80±{\scriptsize0.73}\\
        \midrule
        
        CLIP & 1.12±{\scriptsize0.16} & 17.93±{\scriptsize1.21} \\
        \midrule
        
        SimCLR & 1.44±{\scriptsize0.45} & 14.86±{\scriptsize0.29} \\
        
        SwAV & 1.18±{\scriptsize0.26} & 9.45±{\scriptsize0.61} \\
        
        Barlow Twins & 1.10±{\scriptsize0.10} & 15.26±{\scriptsize0.67} \\
    
        \bottomrule
    \end{tabular}
    
    \caption{
    \textbf{Co$^2$L with two-Stage training.} Accuracy increases largely when the classifier is trained for 30 epochs instead of just one.
    }
    \label{tab:accu_co2l_twostage}
\end{table}

\section{Co$^2$L with Two-Stage Training}
In Table~\ref{tab:accu_co2l_twostage}, after Co$^2$L learns the feature in the streaming phase, instead of training the classifier for one epoch, we train it for 30 epochs. By doing so, accuracy increases significantly. 

\end{document}